\documentclass[10pt, twocolumn, letterpaper]{article}
\pdfoutput=1

\usepackage{cvpr}
\usepackage{times}
\usepackage{epsfig}
\usepackage{graphicx}
\usepackage{amsmath,amssymb}
\usepackage{subfigure}

\cvprfinalcopy

\ifcvprfinal\fi

\usepackage{ifthen}
\newboolean{arxiv}
\setboolean{arxiv}{true}

\def\subsubsection{\paragraph}

\begin{document}

\title{Geometric Polynomial Constraints in Higher-Order Graph Matching} % Replace with your title

\author{Mayank Bansal$^{1,2}$\\ 
$^1$Center for Vision Technologies, SRI International\\
Princeton, NJ, USA\\
{\tt\small mayank.bansal@sri.com}
\and Kostas Daniilidis$^2$\thanks{The authors are grateful for support through the following grants: NSF-IIP-0742304, NSF-OIA-1028009, ARL MAST-CTA W911NF-08-2-0004, ARL RCTA W911NF-10-2-0016, NSF-DGE-0966142, and NSF-IIS-1317788.}\\
$^2$GRASP Lab, University of Pennsylvania\\
Philadelphia, PA, USA\\
{\tt\small kostas@cis.upenn.edu}}

\graphicspath{{figs/}}

\maketitle

\begin{abstract}
Correspondence is a ubiquitous problem in computer vision and graph matching has been a natural way to formalize correspondence as an optimization problem. Recently, graph matching solvers have included higher-order terms representing affinities beyond the unary and pairwise level. Such higher-order terms have a particular appeal for geometric constraints that include three or more correspondences like the PnP 2D-3D pose problems. In this paper, we address the problem of finding correspondences in the absence of unary or pairwise constraints as it emerges in problems where unary appearance similarity like SIFT matches is not available. Current higher order matching approaches have targeted problems where higher order affinity can simply be formulated as a difference of invariances such as lengths, angles, or cross-ratios. In this paper, we present a method of how to apply geometric constraints modeled as polynomial equation systems. As opposed to RANSAC where such systems have to be solved and then tested for inlier hypotheses, our constraints are derived as a single affinity weight based on $n>2$ hypothesized correspondences without solving the polynomial system.  Since the result is directly a correspondence without a transformation model, our approach supports correspondence matching in the presence of multiple geometric transforms like articulated motions.
\end{abstract}

\section{Introduction}
\label{sec:intro}

Graph Matching has been the standard way to formalize correspondence finding as an optimization problem.
It has been employed for matching features between image pairs using appearance and geometric constraints, and for matching 3D point sets using geometric constraints. 
More recent work has extended the graph matching framework to higher-order graphs where affinities representing a match between tuples of features can easily be written as differences of lengths, angles, or cross-ratios.
 As an example, for the problem of matching features between two images, one can directly use a higher-order (hyper) edge of degree $3$ to represent the similarity between a pair of triangles. However, expression of these higher order constraints relies on invariant features that can be computed on each image independently. Thus, one computes a feature vector $f_{i,j,k}$ representing some invariant property measured from the triangle $ijk$ in image $\mathcal{I}$ and feature vector $f_{i',j',k'}$ representing the same property for triangle $i'j'k'$ in image $\mathcal{I'}$. Then, the affinity of the hyper-edge $(i,i',j,j',k,k')$ is measured as a function of the distance $||f_{i,j,k}-f_{i',j',k'}||$.  Several examples of such constraints were shown by Duchenne \etal~\cite{duchenne2011tensor}  like the perspective-invariant feature vector composed of the three cross-ratios measured from each triangle or difference of angles in \cite{park2013fast}. These measures are thus limited to intra-image features which can be computed for each triangle in an image independently and then compared against similar features computed for a triangle in the second image. This limits the applicability of the higher-order matching framework to setups in which such invariant features can be designed. 

In this paper, we propose to extend the idea of higher-order constraints to inter-image constraints of the kind encountered in geometric correspondence problems. As a concrete example, consider the Perspective-$n$-Point (P$n$P) problem where we are given a set of 3D points and their 2D projections (with unknown correspondence) and we are interested in solving for the correspondence using purely geometric constraints. It is well known that there are no geometric invariants between a set of 3D points and their 2D projections. Therefore, one cannot employ higher-order constraints for this problem using the framework proposed in the literature. However, the constraints in this problem setup occur in the form of geometric constraints between corresponding points when one considers a minimal configuration of three point correspondences. Given a 3D triangle to 2D triangle correspondence, one can solve for upto four possible solutions for the P3P problem. However, the existence of a solution is not sufficient to say if the chosen triangle correspondence is correct -- a fourth correspondence is required to verify the solution. The key idea in this paper is to estimate an affinity measure for a $4^{th}$ order hyper-edge directly without first solving the minimal geometric problem completely. To achieve this, we express the geometric problem for each subset of points in a minimal configuration using a univariate polynomial equation. Then, two such minimal sets are consistent only if their polynomial equations share a common root. Thus, the affinity measure can be expressed as a {\em resultant} of the polynomial pair (see section \ref{sec:poly}). 

We believe that with this paper, we advance the state of the art in the following directions:
%\textcolor[rgb]{0,0,1}{Question: Why can't we use the overdetermined solution from four points directly to estimate the affinity value? }
\begin{itemize}
\item
We introduce the novel idea of using the magnitude of the resultant of a pair of polynomial equations as a measure of agreement between the models represented by the equations that can represent a higher-order affinity across graphs.
\item
We solve the correspondence problem based purely on geometric polynomial constraints within an existing graph matching optimization framework.
\item 
As opposed to RANSAC which filters initial correspondences, we start with complete lack of any feature matches. 
\item
As opposed to EM, we enable correspondence finding without computing the underlying geometric transformations, thus enabling matching in the presence of multiple transformations like articulated motion.
\end{itemize}
This is a theoretical contribution which we tested on synthetic data. 
We will provide tests on real data as soon as we establish a parallel implementation for the computation of the affinities which takes the most significant part of the running time.

\subsection{Related Work}
\label{sec:related}

Leordeanu and Hebert~\cite{leordeanu2005spectral} consider the quadratic assignment problem where distances between pairs of features from two images are used to create an affinity matrix and an efficient spectral solution to solving this problem is proposed. Cour \etal~\cite{cour2006balanced} generalize their approach to allow incorporation of additional affine constraints. Schellewald and Schn{\"o}rr~\cite{schellewald2005probabilistic} address the same problem in a convex optimization framework by relaxing the discrete problem into a semidefinite program (SDP). More recently, Zhou and De la Torre~\cite{zhou2013deformable,zhou2012factorized} proposed deformable graph matching (DGM) for matching graphs subject to global rigid and non-rigid geometric constraints. However, they also restrict the choices of transformation to certain classes like similarity, affine and RBF non-rigid and work in the context of image to image matching.

The use of higher-order matching in the computer vision literature has focused on inclusion of constraints derived from higher-order geometric invariants like angles of triangles, cross-ratio along lines etc. This allows more robust matching between features in two images (under affine or plane projective assumptions) or between 3D point clouds. However, the case of matching between 3D and 2D features has not been addressed due to the lack of any geometric invariants between them. Ochs and Brox~\cite{ochs2012higher} apply spectral clustering on a projected hypergraph computed from higher-order tuples of motion trajectories. Using affinities beyond just pairs of trajectories allows them to handle non-translational motion like rotation and scaling.

Many recent approaches have proposed algorithms for computing an assignment matrix given a higher-order graph encapsulating relations between tuples of features. Zass and Shashua~\cite{zass2008probabilistic} approached the hyper-graph matching problem in a probabilistic setting but used certain independence assumptions to factor the model into first-order interactions. Lee \etal~\cite{lee2011hyper} proposed a random-walk approach for higher-order graph matching. Duchenne \etal~\cite{duchenne2011tensor} proposed an extension of the spectral power iteration method for matrix eigen-value problems to tensors and show how it can be used to solve assignment problems on higher-order graphs by expressing the hyper-edge affinities as a tensor. 
 
A number of recent approaches have focused on the computational aspect of the higher-order matching problem. Park \etal~\cite{park2013fast} recently proposed the Higher Order FAst Spectral graph Matching (HOFASM) algorithm that approximates the affinity tensor used for higher-order graph matching resulting in lower memory and computational requirements. However, to exploit the redundancy in the affinity tensor, they require the existence of many tuples of feature points whose corresponding angles are very close to each other. Thus, their approach doesn't directly apply to problems where the same kind of features are not being matched (e.g. 3D to 2D). Cheng \etal~\cite{cheng2013supermatching} also focus on the computational aspects of higher-order matching by defining a compact affinity tensor, devising a sampling strategy to reduce redundancy and optimizing the power iteration method for computational efficiency. While some aspects of their approach are applicable to our formulation, in this paper our focus is on proposing a theoretical framework to allow inclusion of geometric constraints into higher-order problems which lack geometric invariants.

\section{Tensor Matching for Geometric Problems}
\label{sec:tech}

%\subsection{Preliminaries}

\subsection{Tensor formulation for higher-order graph matching}\label{sec:tensors}
Duchenne \etal\cite{duchenne2011tensor} introduced the idea of using a tensor representation for higher-order graph matching problems and we present a brief review of their algorithm in this section.

Consider the problem of matching $N$ points in image $\mathcal{I}$ against $N'$ points in image $\mathcal{I'}$. This problem is equivalent to determining an $N \times N'$ assignment matrix $X$ such that $X_{i,i'}$ is $1$ when point $P_{i}$ is matched to the point $P_{i'}$ and $0$ otherwise. There are additional constraints on the matrix $X$ in the form of unit row or column sums depending on whether we allow many-to-one and one-to-many matching. Given an affinity matrix $H$ such that $H_{i,i'}$ corresponds to the similarity between points $P_i$ and $P_{i'}$, the matching problem can be formulated as the maximization of the cost function given by $\textrm{score}(X) = \sum_{i,i'}{H_{i,i'}X_{i,i'}}$ subject to the row/column stochasticity constraints on $X$. The affinity matrix can be generalized to an affinity tensor such that $H_{i,i',j,j',k,k'}$ represents the similarity between tuples of points $(P_i,P_j,P_k)$ and $(P_{i'},P_{j'},P_{k'})$. In this case the matching cost function can be written as:
\begin{equation}
\textrm{score}(X) = \sum_{i,i',j,j',k,k'}{H_{i,i',j,j',k,k'}X_{i,i'}X_{j,j'}X_{k,k'}}
\label{eq:tensorscore}
\end{equation}

The affinity tensor $H$ in this case is assumed to be a $6$-dimensional super-symmetric tensor. The score function in \eqref{eq:tensorscore} can be written in the tensor notation as:
\begin{equation}
\textrm{score}(\tilde{X}) = \tilde{H} \otimes_3 \tilde{X} \otimes_2 \tilde{X} \otimes_1 \tilde{X}
\label{eq:tensorscore2}
\end{equation}

where $\tilde{X} = vec(X)$ is the $NN'$ vector obtained by concatenating the columns of $X$. $\tilde{H}$ is the tensor form of $H$ of size $(NN')^d$ with $d=3$ in this case.

The score function in \eqref{eq:tensorscore2} can be optimized using the tensor power iteration algorithm proposed by Duchenne \etal~\cite{duchenne2011tensor} which generalizes the idea of power iterations for eigenvalue problems. They also show a version of the algorithm that uses $\ell^1$-norm constraints on the rows of the assignment matrix $X$ to generate a close to boolean output matrix. For further details about these two algorithms, the reader is referred to \cite{duchenne2011tensor}. 
%In this paper, we have employed the tensor power-iteration solver but our approach can be applied as part of the solver proposed by Park \etal~\cite{park2013fast} as well.

%\subsubsection{Building the tensor}
%
%Tensor size $(N_1 N_2)^d$.
%
%$n = \textrm{max}\{N_1,N_2\}$.

\def\Pa{\mathcal{P}}
\def\Pb{\mathcal{P}'}

\subsection{Geometric Constraints for Higher-Order Graphs}\label{sec:geocons}
%In this section, we define the ``Geometric Hyper-graph'' which encodes the geometric constraints in a given problem as weights on the hyper-edges. A higher-order graph-matching framework is then applied on this graph to derive a correspondence set that maximizes the satisfaction of these geometric constraints.

\def\boldtau{{\tau}}

Consider a geometric matching problem where the objective is to match a set $\Pa$ of $n$ points in instance I against a set $\Pb$ of $n'$ points in instance II, subject to some geometric constraints. We will assume that there are no appearance constraints that can be used to aid correspondence. Additionally, we will assume that there are no geometric invariants that can be computed and matched between subsets of points in $\Pa$ or $\Pb$. This is the case, for example, in the ``Perspective-$n$-Point'' (P$n$P) problem where $\Pa$ will specify a set of 3D points and $\Pb$ will specify the corresponding set of 2D projections of these points into a calibrated camera. In a problem like this, the geometric constraints are typically specified for a ``minimal configuration'' of $m$ point correspondences from the sets $\Pa$ and $\Pb$. For concreteness, let $S = \{(p_1,p'_1),(p_2,p'_2),\ldots,(p_m,p'_m)\}$ be a set of $m$ correspondences from the set $\Pa \times \Pb$. Without loss of generality, let $\mathcal{F}_\tau(p_i,p'_i) = 0$ be a (polynomial) function specifying the geometric constraint between points $p_i$ and $p'_i$ with the unknown parameter vector $\tau$ providing a parametrization of the global geometric configuration. The set of constraints $\mathcal{C} \equiv \{\mathcal{F}_\tau(p_i,p'_i) = 0\}$ for $i = 1,\ldots,m$ can be used jointly to solve for the parameter vector $\boldtau$ by successive elimination of variables to reduce the set $\mathcal{C}$ to a single univariate polynomial equation of (say) degree $k$:
\begin{equation}
q_S(x) \equiv a_{S,k}x^k + a_{S,{k-1}}x^{k-1} + \ldots + a_{S,0} = 0
\label{eq:qS}
\end{equation}

where the subscript $S$ reflects the dependence of the polynomial on the chosen minimal set $S$. In this paper, we focus on the problems for which such a reduction to a univariate polynomial is possible and we explicitly derive the formulation for three common problems in computer vision.

\begin{figure}[t]
	\centering
		\includegraphics[width=0.5\textwidth,clip=true,trim=0 368 75 0]{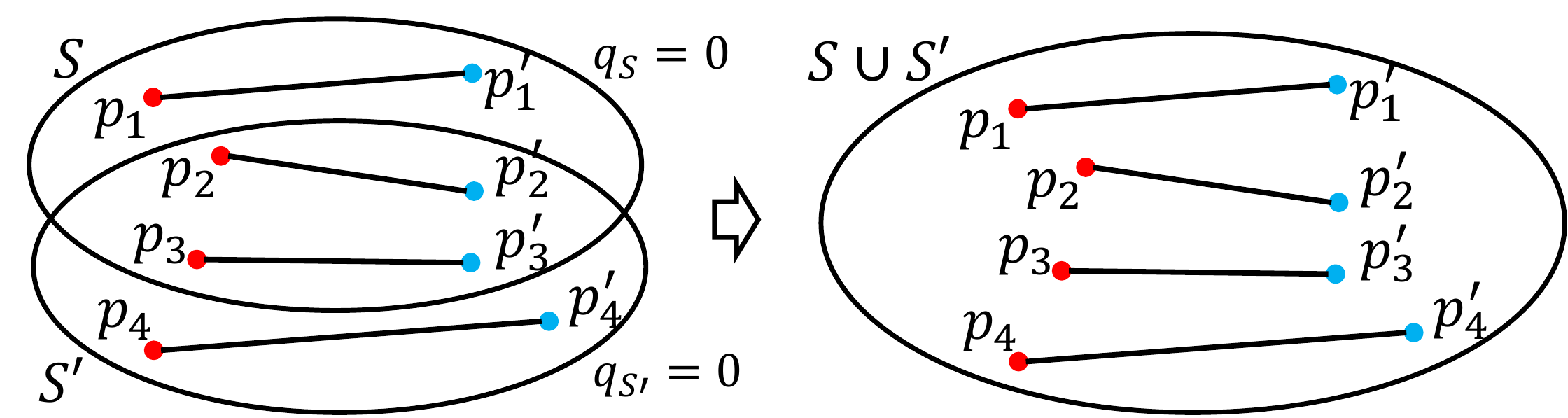}
		\caption{Correspondences in sets $S$ and $S'$ are the hyper-edges corresponding to the minimal configuration and generate constraints in the form of polynomial equations $q_S=0$ and $q_{S'}=0$ respectively. These sets are combined to form a new hyper-edge with weight given by the resultant of the Sylvester matrix $M(q_S,q_{S'})$ of the two polynomials.}
		\label{fig:ill}
\end{figure}

The polynomial equation \eqref{eq:qS} can be solved to obtain solution(s) for the parameter $x$ and then back-substituted into the set $\mathcal{C}$ to determine the full parameter vector $\tau$. However, this is not sufficient to associate a cost with the set $S$ \footnote{For certain problems, the non-existence of real solutions for the equation $q_S(x)$ can be used to associate a cost value. However, there are robustness issues with this approach if the data points are noisy.}. Typically, a new point correspondence $(p_{m+1},p'_{m+1})$ is used to test and validate the solutions of \eqref{eq:qS}. However, this approach needs one to solve for the full parameter vector $\tau$ and then evaluate the new correspondence. Instead, we propose to use the ${m+1}^{th}$ correspondence to define another minimal set $S'$ that shares $m-1$ correspondences with the set $S$ and define a cost directly using the polynomial equations $q_S$ and $q_{S'}$. This set $S' = \{(p_2,p'_2),(p_3,p'_3),\ldots,(p_{m+1},p'_{m+1})\}$ is obtained from $S$ by replacing the first correspondence $(p_1,p'_1)$ by another correspondence $(p_{m+1},p'_{m+1})$. Since this is a minimal set, we can derive a polynomial equation $q_{S'}(x)$ where $x$ is the same variable\footnote{Note that problem-specific tricks might be required to ensure that the variable $x$ is indeed a variable that can be shared between the polynomials $q_S$ and $q_{S'}$.} as in the polynomial for set $S$. The set $S \cup S'$ consists of $m+1$ correspondences and is geometrically consistent if the polynomials $q_S$ and $q_{S'}$ share a common root. In section \ref{sec:poly}, we describe an approach to quantify the existence of a common root between the two univariate polynomials using the Sylvester resultant. Using the resultant, we directly measure the distance of the two polynomials from co-primeness without needing to evaluate their roots and solving for the full parameter vector $\tau$. Fig.~\ref{fig:ill} illustrates this construction for $m=3$ which is the minimal configuration for the $P3P$ problem discussed in section \ref{sec:p3p}.

\subsection{Polynomial Resultant as Edge Affinities}\label{sec:poly}
Consider the family of univariate degree-$n$ polynomial equations $\{p_i(x)\}$ defined as:
\begin{equation}
p_i(x) \equiv a_{i,n}x^n + a_{i,{n-1}}x^{n-1} + \ldots + a_{i,0} = 0
\label{eq:pi}
\end{equation}

The polynomials are assumed to have a unitary $2$-norm i.e.
\begin{equation}
||(a_{i,n},a_{i,n-1},\ldots,a_{i,0})^T||_2 = 1
\end{equation}

Given two polynomials $p_i$ and $p_j$ from this family, we are interested in the problem of determining if they have a common root. This problem can be approached by considering the resultant matrix of the polynomials. A resultant matrix of two polynomials is a matrix obtained from the polynomial coefficients with the property that the polynomials have a common root if and only if the matrix has a zero determinant. Two of the most common resultant matrices often employed are the {\em Sylvester} and the {\em B$\acute{e}$zout} matrices. In this paper, we employ the resultant of the Sylvester matrix corresponding to the polynomial pair $(p_i, p_j)$ because of its simpler form in terms of the polynomial coefficients in comparison to the B$\acute{e}$zout matrix. The Sylvester matrix $M(p_i, p_j)$ is defined as follows:

\ifthenelse {\boolean{arxiv}}
{
\begin{align}
& M(p_i,p_j) =  \\
& \left(\begin{array}{ccccccc}
a_{i,n} & a_{i,n-1} & \ldots    & a_{i,0} &          	&   & 			\\
        & a_{i,n}   & a_{i,n-1} & \ldots  & a_{i,0}   &       &  			\\
				&           & \ddots    & \ddots  & \ddots    & \ddots       &  \\
        &           &           & a_{i,n} & a_{i,n-1} & \ldots & a_{i,0}\\
\hline
a_{j,n} & a_{j,n-1} & \ldots    & a_{j,0} &          &  & 			\\
      & a_{j,n}   & a_{j,n-1} & \ldots  & a_{j,0}   &       &  			\\
  &           & \ddots    & \ddots  & \ddots    & \ddots       & \\
       &     &          & a_{j,n} & a_{j,n-1} & \ldots & a_{j,0}\\
\end{array}\right)_{2n\times 2n}
\label{eq:Mpipj}
\end{align}
}
{
\begin{equation}
M(p_i,p_j) = 
\left(\begin{array}{ccccccc}
a_{i,n} & a_{i,n-1} & \ldots    & a_{i,0} &          	&   & 			\\
        & a_{i,n}   & a_{i,n-1} & \ldots  & a_{i,0}   &       &  			\\
				&           & \ddots    & \ddots  & \ddots    & \ddots       &  \\
        &           &           & a_{i,n} & a_{i,n-1} & \ldots & a_{i,0}\\
\hline
a_{j,n} & a_{j,n-1} & \ldots    & a_{j,0} &          &  & 			\\
      & a_{j,n}   & a_{j,n-1} & \ldots  & a_{j,0}   &       &  			\\
  &           & \ddots    & \ddots  & \ddots    & \ddots       & \\
       &     &          & a_{j,n} & a_{j,n-1} & \ldots & a_{j,0}\\
\end{array}\right)_{2n\times 2n}
\label{eq:Mpipj}
\end{equation}
}

It is well known that the Sylvester matrix is a resultant matrix \cite{laidacker1969another}. Given a vector $ \textbf{y} = (x^{2n-1}, x^{2n-2}, \ldots, x, 1)^T $, the system $M\textbf{y} = 0$ has a solution if and only if the polynomials $p_i(x)$ and $p_j(x)$ have a common root. This implies that the square matrix $M$ has to be rank deficient for the polynomials to share a root. Thus, we can use the magnitude of the smallest singular value $\sigma_{min}(M)$ of $M$ as a measure of co-primeness of the polynomials $p_i$ and $p_j$. In addition, the last non-zero row of the $R$ matrix obtained by a $QR$-factorization of the matrix $M$ represents the coefficients of the GCD polynomial for $p_i$ and $p_j$ (see \cite{corless2004qr}). Because of this property, we can also use the absolute value of the last element $R_{2n,2n}$ of $R$ to measure the co-primeness of $p_i$ and $p_j$. 

The above formulation allows us to assign an affinity value to the hyper-edge $S \cup S'$ defined in section \ref{sec:geocons} as follows:
\begin{equation}
H_{S \cup S'} = e^{-|M(q_S,q_{S'})|/\rho}
\label{eq:affinity}
\end{equation}

where $|M(q_S,q_{S'})|$ represents the resultant value estimated either from the SVD as $\sigma_{min}(M(q_S,q_{S'}))$ or from the QR factorization as $|R_{2n,2n}|$, and $\rho$ is a parameter that determines the spread of the exponential function in \eqref{eq:affinity}. 

For the simulations in this paper, we experimented with both methods for computing the resultant value and found them to perform equally well. Therefore, we picked the $QR$ factorization based approach for all the experimental results in this paper because of its lower computational complexity as compared to SVD. The parameter $\rho$ was set empirically in our experiments and kept constant for all the instances and problems.

%\begin{equation}
%y = 
%\left(\begin{array}{c}
%x^{2n-1} \\ x^{2n-2} \\ \vdots \\ x \\ 1
%\end{array}\right)
%\label{eq:}
%\end{equation}

%\begin{equation}
%My = 0
%\label{eq:}
%\end{equation}

\subsubsection{Overall Algorithm.} Collecting the ideas presented in the previous sections, we propose the following general framework to approach graph matching for geometric problems.
\begin{enumerate}
	\item Consider the correspondence problem in its minimal configuration (of size say $m$) and {\em analytically} derive the minimal geometry equations relating the $m$ pairs of observations (points) to the unknown geometric variables.
	\item {\em Analytically} combine the constraint equations over the minimal set into a univariate polynomial with coefficients dependent on the $m$ pairs of observations in the minimal set.
	\item For a given problem instance:
	\begin{enumerate}
		\item Sample hyper-edge tuples of ${m+1}^{th}$ order from the space of all possible ${m+1}^{th}$ order tuples of correspondences. 
		\item For each sample, {\em numerically} compute the coefficients of the two univariate polynomials derived in step (2) above. Compute the affinity value for this hyper-edge by plugging-in these coefficients into equation \eqref{eq:affinity}.
		\item Apply the tensor power iteration method from \cite{duchenne2011tensor} to the computed affinity tensor to compute an assignment matrix.
	\end{enumerate}

\end{enumerate}

\section{Formulation for Specific Problems}
In the following, we describe novel formulations for three specific geometric problems. Two of the problems deal with absolute camera pose recovery given 3D points as $\Pa$ and their 2D projections as $\Pb$. The third problem deals with the relative camera pose problem (visual odometry) in a setting where the camera rotation axis is known (or can be estimated by a directional correspondence). In this case, only three point correspondences are required and we show how we can formulate this in our graph-matching framework. The sets $\Pa$ and $\Pb$ in this case are both 2D image points.

\subsection{Three-point Calibrated Absolute Pose Problem (P3P)}\label{sec:p3p}

\def\xa{X_a}
\def\xb{X_b}
\def\xc{X_c}
\def\xd{X_d}
\def\ua{u_a}
\def\ub{u_b}
\def\uc{u_c}
\def\ud{u_d}

There are several algorithms for this classic problem in the literature. However, in this paper we will work with the formulation proposed by Fischler and Bolles~\cite{fischler1981random} in their classic RANSAC paper. 

\subsubsection{Minimal Setup.}
The minimal setup consists of three 3D points $\xa$, $\xb$ and $\xc$ being observed by a camera at a 3D location $O$. Given image projections $\ua$, $\ub$ and $\uc$ of these points and the camera calibration matrix $K$, the problem is to estimate the absolute camera pose. Given the pairwise distances between the 3D points i.e. $||\xa - \xb|| = R_{ab}, ||\xb - \xc|| = R_{bc}, ||\xc - \xa|| = R_{ca}$ and the angle subtended by each pair of image points at the camera center i.e. $\textrm{cos}(\angle\xa O\xb) = C_{ab}, \textrm{cos}(\angle\xb O\xc) = C_{bc}, \textrm{cos}(\angle\xc O\xa) = C_{ca}$, the problem is to estimate the distances of the points from the camera center. Let these unknowns be denoted by $a$, $b$ and $c$, i.e. $||\xa - O|| = a, ||\xb - O|| = b, ||\xc - O|| = c$. 

Each of the triangles $OAB$, $OBC$ and $OCA$ provides a quadratic constraint per the cosine law:
\begin{align}
R_{ab}^2 &= a^2 + b^2 - 2ab C_{ab} \label{eq:quadAB}\\
R_{bc}^2 &= b^2 + c^2 - 2bc C_{bc} \label{eq:quadBC}\\
R_{ca}^2 &= c^2 + a^2 - 2ca C_{ca} \label{eq:quadCA}
\end{align}

By introducing new variables $x$ and $y$ such that $x = b/a$ and $y = c/a$, algebraic manipulation of the system \eqref{eq:quadAB}-\eqref{eq:quadCA} leads to the following quartic polynomial equation in $x$:
\begin{equation}
G^{abc}_4 x^4 + G^{abc}_3 x^3 + G^{abc}_2 x^2 + G^{abc}_1 x + G^{abc}_0 = 0
\label{eq:p3pabc}
\end{equation}

Solving this polynomial equations leads to four possible solutions for the variable $x$ which is the ratio of the depths of the points $\xb$ and $\xa$ from the camera center $O$. The depths $a$, $b$ and $c$ can then be solved for using the system \eqref{eq:quadAB}-\eqref{eq:quadCA}. 

\subsubsection{Hyper-edge.}
As outlined in section \ref{sec:geocons}, since the minimal solution may not provide any constraints in the general case, we add another 3D point $\xd$ (and the corresponding image point $\ud$) and derive another quartic polynomial by considering the tetrahedron $OABD$. With $||\xb-\xd|| = R_{bd}, ||\xd - \xa|| = R_{da}$, $\textrm{cos}(\angle\xb O\xd) = C_{bd}, \textrm{cos}(\angle\xd O\xa) = C_{da}$ and additional variable $||\xd - O|| = d$, we get the following quartic polynomial:

\begin{equation}
G^{abd}_4 x^4 + G^{abd}_3 x^3 + G^{abd}_2 x^2 + G^{abd}_1 x + G^{abd}_0 = 0
\label{eq:p3pabd}
\end{equation}

where $x$ is defined again as $x = b/a$ and $z$ is defined as $z = d/a$.

We can now define a $4^{th}$ order hyper-edge $e \equiv (\xa,\ua,\xb,\ub,\xc,\uc,\xd,\ud)$ with the edge affinity $H_e$ defined by \eqref{eq:affinity} with $S = (\xa,\ua,\xb,\ub,\xc,\uc)$ and $S' = (\xa,\ua,\xb,\ub,\xd,\ud)$; $M(q_S,q_S')$ is the $8\times 8$ Sylvester matrix of the two polynomials \eqref{eq:p3pabc} and \eqref{eq:p3pabd}. Note that this formulation works since the variable $x = b/a$ is shared between the two equations and hence the problem of looking for a common root for the polynomial pair is well defined.

%Similarly, triangle ABD leads to:
%\begin{align}
%||\xa-\xb|| = R_{ab}, ||\xb-\xd|| = R_{bd}, ||\xd - \xa|| = R_{da}\\
%\textrm{cos}(\angle\xa O\xb) = C_{ab}, \textrm{cos}(\angle\xb O\xd) = C_{bd}, \textrm{cos}(\angle\xd O\xa) = C_{da}\\
%||\xa - O|| = a, ||\xb - O|| = b, ||\xd - O|| = d
%\label{eq:}
%\end{align}

%Introducing new variables $x$ and $z$ such that $x = b/a$ and $z = d/a$, we get another quartic equation:

\ifthenelse {\boolean{arxiv}}
{
\begin{figure*}[t]
}
{
\begin{figure}[t]
}
	\centering
		\begin{minipage}[c]{5mm}(a)\end{minipage} \hfill
		\begin{minipage}[c]{0.34\textwidth}{\includegraphics[width=\textwidth]{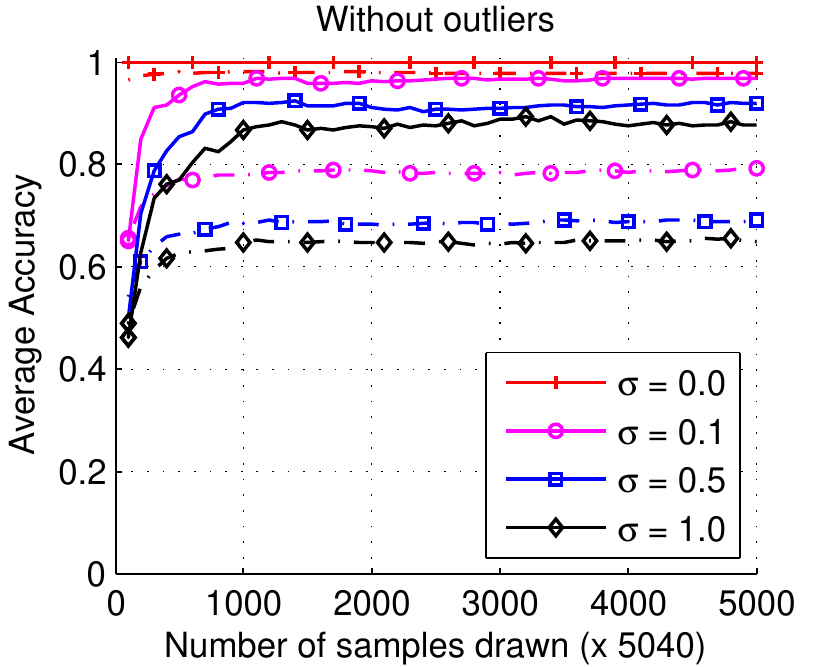}}\end{minipage} \hfill
		\begin{minipage}[c]{5mm}(b)\end{minipage} \hfill
		\begin{minipage}[c]{0.525\textwidth}{\includegraphics[width=\textwidth,clip=true,trim=0 0 25 0]{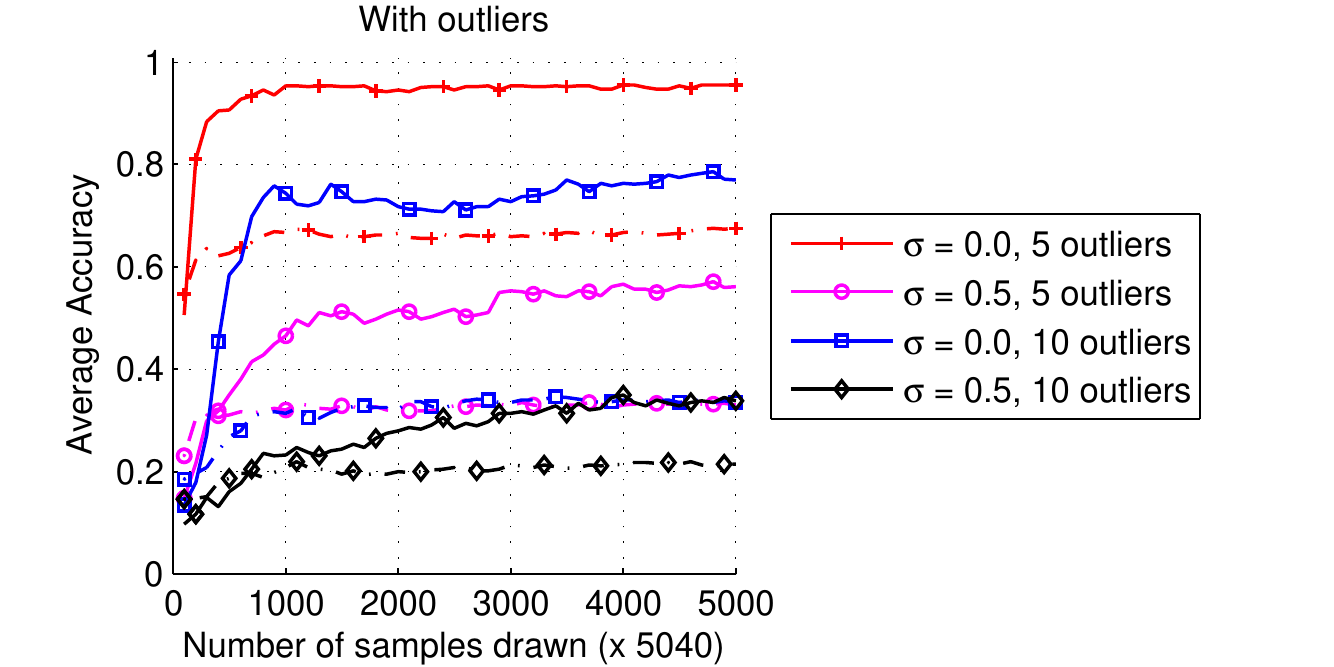}}\end{minipage}
		\caption{\textbf{Simulation results for the P3P Problem.} Average matching accuracy across $100$ randomly generated instances of the P3P problem ($n = 10$) for different amounts of Gaussian noise. The {\em solid} and {\em dashed} curves represent the performance of tensor power-iteration algorithm \cite{duchenne2011tensor} with and without the $\ell^1$-norm constraint on the assignment matrix respectively. In (a) no outliers were added; in (b) $50\%$ and $100\%$ additional random image points were added as outliers to the inlier set of $10$ points.}
		\label{fig:p3p}
\ifthenelse {\boolean{arxiv}}
{
\end{figure*}
}
{		
\end{figure}
}

\subsection{Three-plus-One Calibrated Relative Pose Problem (3P1)}

In this problem setup, recently proposed by Naroditsky \etal~\cite{naroditsky2012two}, we are given three image correspondences $q_i \leftrightarrow q'_i, i = 1, 2, 3$ in two calibrated views along with a single directional correspondence $d \leftrightarrow d'$. The problem is to determine the essential matrix $E$ relating the two cameras and thus estimate the relative pose between the cameras. It was shown in \cite{naroditsky2012two} that by taking into account the directional constraint, the degrees of freedom in the essential matrix can be reduced from $5$ to $3$ and the problem can be formulated in closed-form as the solution of a univariate quartic polynomial $\sum_{i=0}^{4}{h_i x^i} = 0$, where the variable $x = \textrm{cos}(\theta)$ corresponds to the cosine of the one-parameter rotation angle between the two cameras. This allows us to formulate this problem in the same manner as the P3P problem using a $4^{th}$ order affinity tensor derived from the resultant of a pair of these quartic polynomials. However, in this case we have multiple choices for the polynomial pairs for each hyper-edge $(q_1,q'_1,q_2,q'_2,q_3,q'_3,q_4,q'_4)$ since the shared variable $x$ is a camera parameter and is not dependent on the points used (unlike the P3P case where it was dependent on the depths of two of the points). Thus, we can derive ${4 \choose 3} = 4$ different polynomial equations from these $4$ correspondences which can be paired to result in ${4 \choose 2} = 6$ different resultant values for the hyper-edge affinity. For the simulations in this paper, we have used one of these $6$ values as the hyper-edge affinity although more complex values derived from multiple of these is also possible. 

\subsection{Two-point Calibrated Absolute Pose for a Vertical Camera (up2p)}
Kukelova \etal~\cite{kukelova2011closed} proposed a closed-form solution to the absolute pose problem for a vertical camera from two 2D-3D correspondences. Since the camera is parameterized by a single angle $\theta$ about the vertical axis, the problem has a total of four unknowns -- the three components of the translation vector and the camera rotation $\theta$. Thus, unlike the P3P case, only two correspondences are required to estimate the camera pose. It was shown in \cite{kukelova2011closed} that the constraints from the minimal set of two correspondences can be combined to derive a univariate quadratic equation $\sum_{i=0}^{2}{h_i x^i} = 0$, where the variable $x = \textrm{tan}(\frac{\theta}{2})$ is a function of the camera rotation and is thus not dependent on the points chosen. Therefore, we can formulate this problem using a $3^{rd}$ order affinity tensor derived from the resultant of a pair of these quadratic equations. In this case, we will get a $4\times 4$ Sylvester matrix from which we can compute the tensor affinity value for each hyper-edge like before. Note that, in this case, we again have ${3 \choose 2} = 3$ different quadratics from the minimal ($3$ correspondence) set from which we can compute ${3 \choose 2} = 3$ different resultant values. The simulations in this paper use only one of these $3$ values to define the hyper-edge affinity.

\section{Experiments}
\label{sec:expt}

We performed simulations for each of three problems to characterize the robustness of our geometric tensor formulation under varying noise and outlier configurations. For each problem simulation, we generated $100$ instances of the problem randomly and used the proposed method to compute an assignment matrix $X$ for each instance. To characterize the dependence of the algorithm accuracy on the number of sampled hyper-edges, we keep track of the affinity tensor $H_i$ for each sample size $s_i$ and compute the corresponding assignment matrix $X_i$ using the tensor $H_i$. The matching accuracy for each sample size $s_i$ is measured as the number of good matches in the computed $X_i$ divided by $n$ where $n$ is number of points used for the specific simulation. In the following, we report results on the accuracy value averaged over the $100$ instances under different simulation conditions and provide details about the specific simulation setup for each problem. We used the authors' \cite{duchenne2011tensor} implementation of the tensor power-iteration algorithm and extended it to allow inclusion of $4^{th}$ order affinities. In all cases, we assume projection to a $640 \times 480$ image from a camera with an effective focal-length of $f_u = f_v = 1000$.

\ifthenelse {\boolean{arxiv}}
{
\begin{figure*}[t]
}
{
\begin{figure}[t]
}
	\centering
		\begin{minipage}[c]{5mm}(a)\end{minipage} \hfill
		\begin{minipage}[c]{0.4\textwidth}{\includegraphics[width=\textwidth]{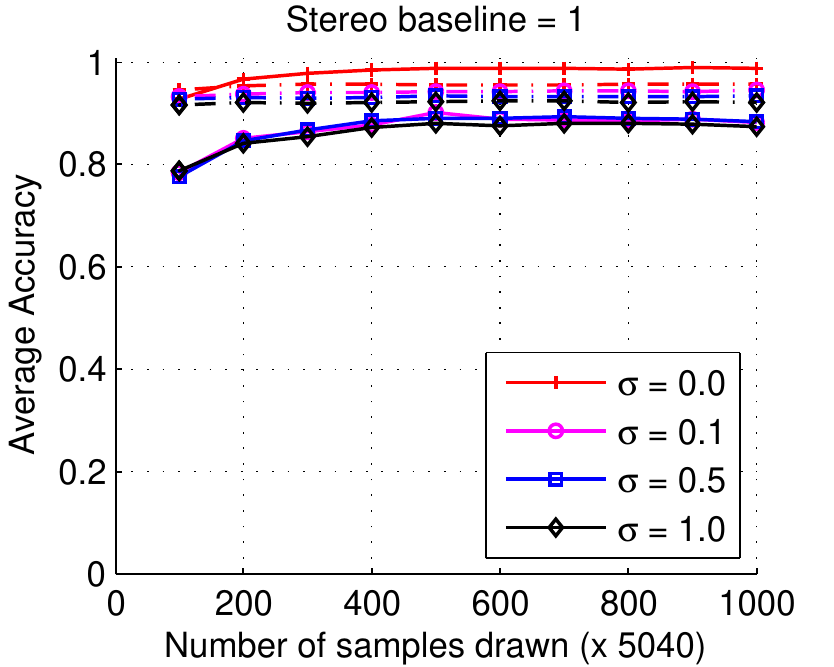}}\end{minipage} \hfill
		\begin{minipage}[c]{5mm}(b)\end{minipage} \hfill
		\begin{minipage}[c]{0.4\textwidth}{\includegraphics[width=\textwidth]{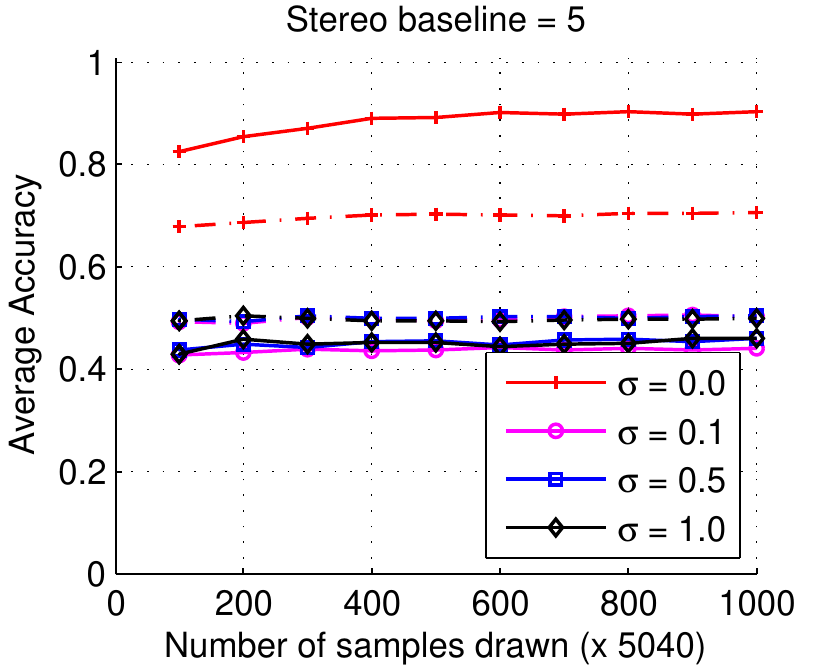}}\end{minipage} \\
		\vspace{5mm}
		
		\begin{minipage}[c]{5mm}(c)\end{minipage} \hfill
		\begin{minipage}[c]{0.4\textwidth}{\includegraphics[width=\textwidth]{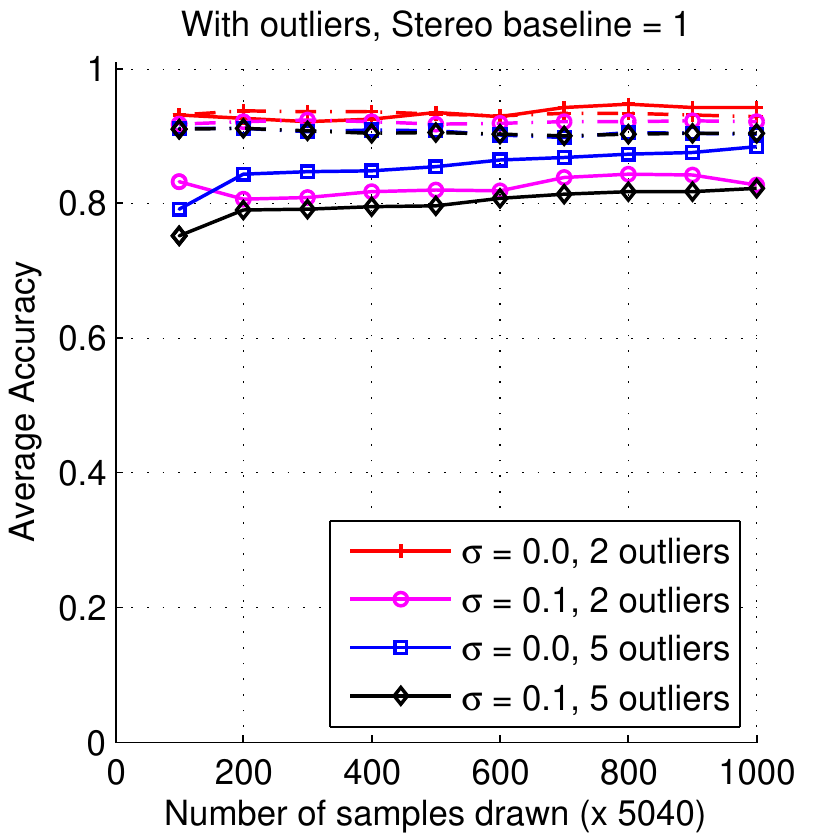}}\end{minipage}
		\begin{minipage}[c]{5mm}(d)\end{minipage} \hfill
		\begin{minipage}[c]{0.4\textwidth}{\includegraphics[width=\textwidth]{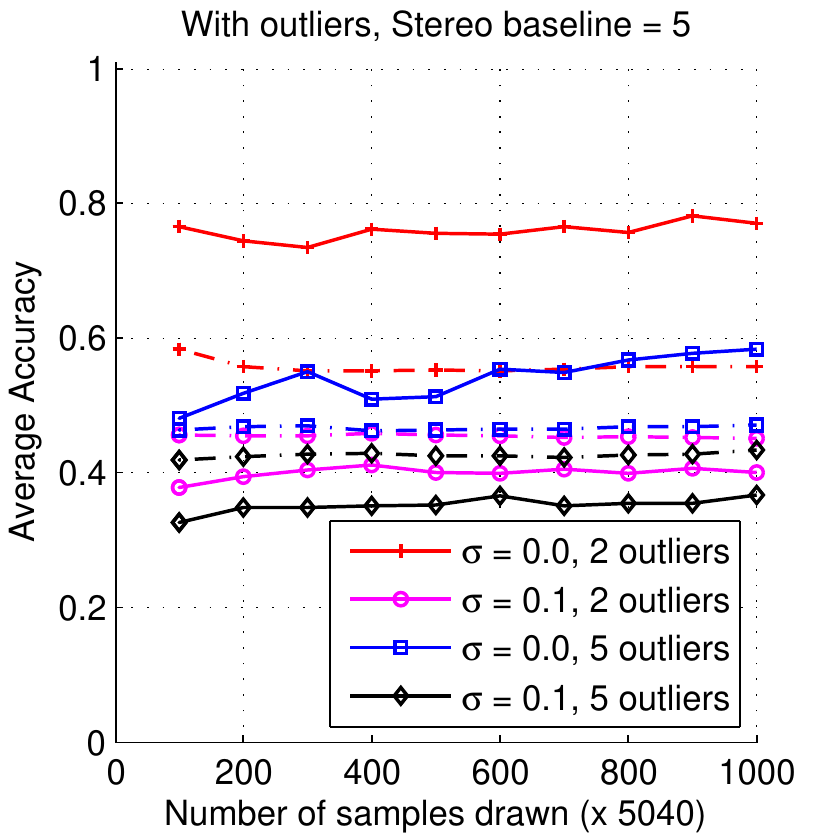}}\end{minipage}
		\caption{\textbf{Simulation results for the 3P1 Problem.} Average matching accuracy across $100$ randomly generated instances of the 3P1 problem ($n = 10$) for different amounts of Gaussian noise. The {\em solid} and {\em dashed} curves represent the {\em sparse} and {\em dense} tensor power-iteration algorithms \cite{duchenne2011tensor}. The left and right columns compare performance for two different values of the baseline between the two cameras. The top and bottom rows show results without and with added outliers respectively.}
		\label{fig:3p1}
\ifthenelse {\boolean{arxiv}}
{
\end{figure*}
}
{		
\end{figure}
}

\subsection{P3P Simulation}

\subsubsection{Simulation Setup.}
For each problem instance, $n=10$ 3D points were generated uniformly at random in a $4\times 4 \times 4$ cube centered at the origin. For each instance, the camera center was also chosen uniformly at random on the surface of a sphere with radius $12$ centered at the origin. The camera rotation was set so that its optical axis passes through the origin ensuring that all the points are always within the camera FOV.

\subsubsection{Noise Sensitivity.}
Fig.~\ref{fig:p3p} compares the accuracy achieved by our algorithm for different amounts of Gaussian noise added to the image points as a function of the number of random tensor edges sampled from the sample space. The solid curves represent the accuracy using the {\em sparse} tensor power-iteration algorithm (i.e. with the $\ell^1$-norm constraint on $X$) while the dashed curves were obtained for the dense formulation. It is clear that the sparse solver performs significantly better than the dense solver. In Fig.~\ref{fig:p3p}\,(a), we show results without any outliers added to the image point set. For the noise-free case $(\sigma = 0)$, we note that the algorithm achieves $100\%$ accuracy in a very small number of samples indicating that our polynomial-based cost is very effective in enforcing the geometric constraints. The other plots correspond to increasing noise level from $\sigma=0.1$ to $\sigma=1.0$ pixel. As expected, noise in the image coordinates interferes with the geometric constraints and hence many more samples are required for the graph-matching to be able to deduce the correct correspondence. Also note that the curves asymptote at less that $100\%$ accuracy because of the inability of purely geometric constraints derived from noisy observations being able to guide the matching process.

\subsubsection{Performance with Outliers.} In Fig.~\ref{fig:p3p}\,(b), we show results for simulations with random outlier points added to the image. Thus, for the $5$ outlier scenario, the simulation generated $n=10$ uniformly random 3D points as before, projected them to the image to generate $10$ 2D points, and then added $5$ additional 2D points uniformly at random to simulate outliers. Then, Gaussian noise with $\sigma$ specified by the experiment was added to all the 2D points. The plots show that the approach can handle a large number of outliers in noise-free case but the performance degrades rapidly when combined with image noise. Also note that in this case the assignment matrix $X$ is non-square and we are therefore solving a subgraph matching problem here.

\subsection{3P1 Simulation}

\ifthenelse {\boolean{arxiv}}
{
\begin{figure*}[t]
}
{
\begin{figure}[t]
}
	\centering
		\begin{minipage}[c]{5mm}(a)\end{minipage} \hfill
		\begin{minipage}[c]{0.34\textwidth}{\includegraphics[width=\textwidth]{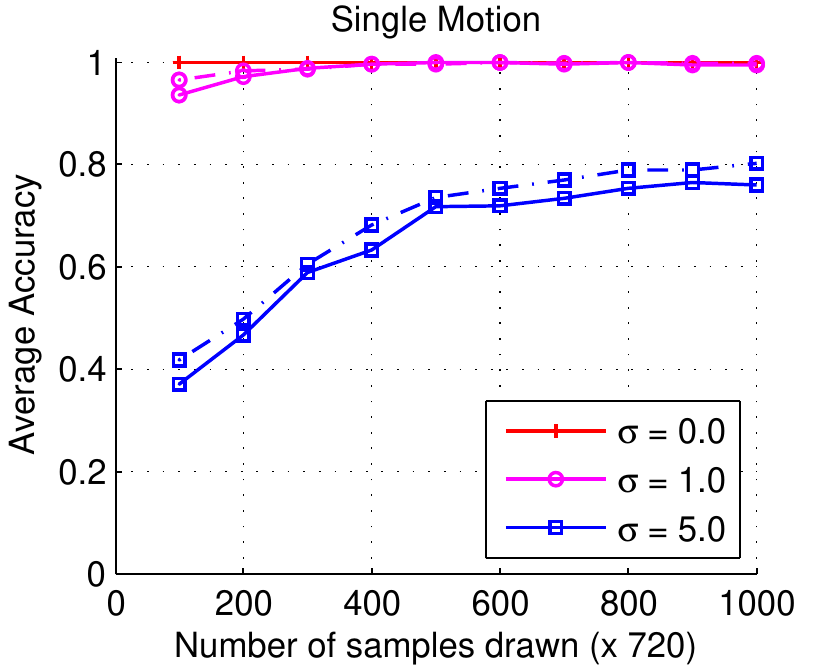}}\end{minipage} \hfill
		\begin{minipage}[c]{5mm}(b)\end{minipage} \hfill
		\begin{minipage}[c]{0.5\textwidth}{\includegraphics[width=\textwidth,clip=true,trim=0 0 25 0]{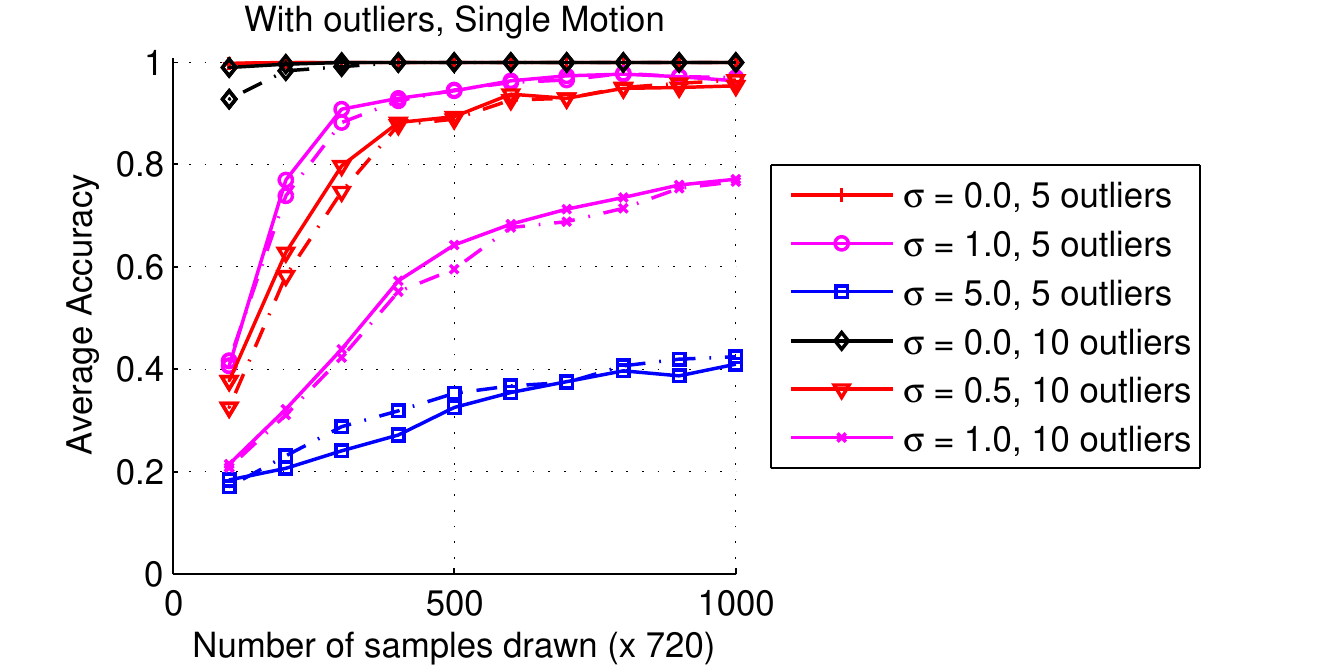}}\end{minipage}\\
		\vspace{5mm}
		\begin{minipage}[c]{5mm}(c)\end{minipage} 
		\begin{minipage}[c]{0.5\textwidth}{\includegraphics[width=\textwidth,clip=true,trim=0 0 25 0]{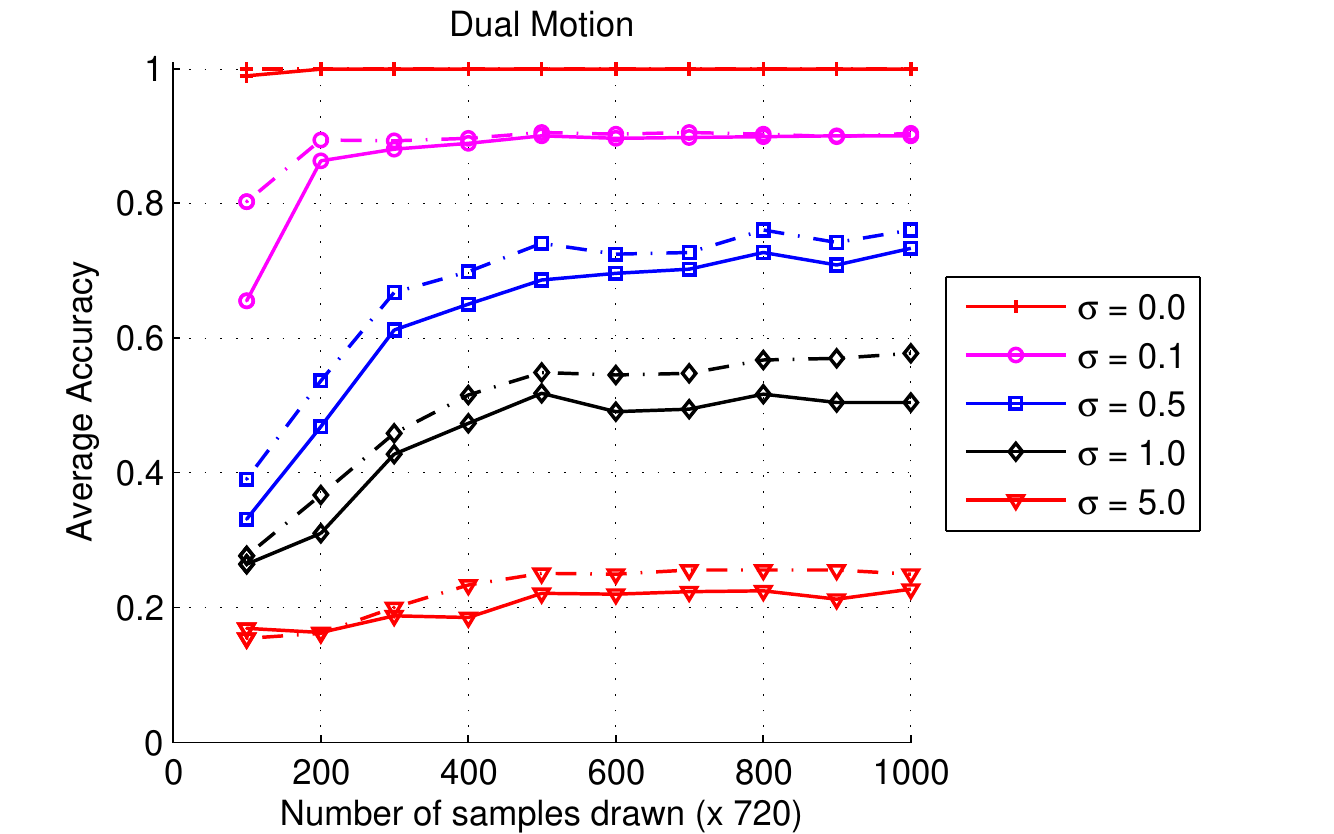}}\end{minipage}
		\caption{\textbf{Simulation results for the Up2p Problem.} Average matching accuracy across $100$ randomly generated instances of the Up2p problem ($n = 10$) for different amounts of Gaussian noise. The {\em solid} and {\em dashed} curves represent the {\em sparse} and {\em dense} tensor power-iteration algorithms \cite{duchenne2011tensor}. In (a) and (b) the data points all followed a single motion; in (c) the $10$ points were split in two sets of $5$ points each following a different motion. In (b), we also show performance with $50\%$ and $100\%$ additional 2D points as outliers for the single motion case.}
		\label{fig:up2p}
\ifthenelse {\boolean{arxiv}}
{
\end{figure*}
}
{		
\end{figure}
}

\subsubsection{Simulation Setup.}
For each problem instance, $n=10$ 3D points were generated uniformly at random in a $4\times 4 \times 4$ cube centered at the origin. For each instance, the first camera center was chosen uniformly at random on the equator of a sphere with radius $12$ centered at the origin. The second camera center was chosen uniformly at random in the equatorial plane at a displacement of $b$ units ({\em stereo baseline}) from the first center. We report results for two values of $b$ in Fig.~\ref{fig:3p1}. The rotation for both cameras were set so that their optical axis passes through the origin ensuring that all the points are always within the camera FOV. Thus, instead of simulating the directional correspondence, we assume that the relative camera rotation is around the vertical axis only.

\subsubsection{Noise Sensitivity.}
Fig.~\ref{fig:3p1} compares the accuracy achieved by our algorithm for different amounts of Gaussian noise added to the image points {\em (in both images)} as a function of the number of random tensor edges sampled from the sample space. Like in the P3P plots, the solid and dashed curves represent the accuracy using the {\em sparse} and {\em dense} tensor power-iteration algorithms respectively with the sparse solver again performing significantly better. Panels (a) and (b) compare the performance for two different baseline settings in the simulation with $b=1$ and $b=5$. We note that for the smaller baseline, the points are less likely to confuse with each other and are thus more robust to noise. For the larger baseline, the performance degrades rapidly with added noise and does not reach $100\%$ accuracy even in the noise-free case indicating likely degenerate conditions for this problem.
 
\subsubsection{Performance with Outliers.} Fig.~\ref{fig:3p1}\,(c) and Fig.~\ref{fig:3p1}\,(d) show results with random outliers added to the second image. The method is quite robust against both outliers and image noise for the shorter baseline but the performance degrades rapidly for the larger baseline.

\subsection{Up2p Simulation}
\subsubsection{Simulation Setup.}
The camera setup in this case was similar to the 3P1 case to ensure a vertical camera with rotation only about the vertical axis. For this problem, we also simulated multiple motions for the results shown in Fig.~\ref{fig:up2p}\,(c). To simulate two motions, for each problem instance, a random rotation about the vertical axis and a random unit translation vector were generated and applied to half of the 3D points before projecting them using the global camera model. Gaussian noise was then added to all the projected points as usual.

\subsubsection{Noise Sensitivity.} Fig.~\ref{fig:up2p}\,(a) shows that for the single motion case, this problem has much better noise handling characteristics than the P3P model. For the dual-motion case in Fig.~\ref{fig:up2p}\,(c), the noise robustness is poorer but the algorithm still shows good performance even with the combination of two motions and image noise.

\subsubsection{Performance with Outliers.} Fig.~\ref{fig:up2p}\,(b) shows results with random outliers added to the image and we note that this method is more robust than the P3P problem in this evaluation as well.
 
%We note that the plots do not depict a monotonic improvement in accuracy with increasing number of samples drawn. \textcolor[rgb]{1,0,0}{This is explained by the fluctuation in the ratio of sampled hyper-edges that contribute to noisy tensor affinity values.}

%\subsection{Numerical Stability with Noise-Free Data}
%
%\subsection{Computational Complexity}
%
%\subsection{Noise Sensitivity}

%Refer to Table~\ref{table:headings}.
%\setlength{\tabcolsep}{4pt}
%\begin{table}
%\begin{center}
%\caption{Font sizes of headings. Table captions should always be
%positioned {\it above} the tables. The final sentence of a table
%caption should end without a full stop}
%\label{table:headings}
%\begin{tabular}{lll}
%\hline\noalign{\smallskip}
%Heading level & Example & Font size and style\\
%\noalign{\smallskip}
%\hline
%\noalign{\smallskip}
%Title (centered)  & {\Large \bf Lecture Notes \dots} & 14 point, bold\\
%1st-level heading & {\large \bf 1 Introduction} & 12 point, bold\\
%2nd-level heading & {\bf 2.1 Printing Area} & 10 point, bold\\
%3rd-level heading & {\bf Headings.} Text follows \dots & 10 point, bold
%\\
%4th-level heading & {\it Remark.} Text follows \dots & 10 point,
%italic\\
%\hline
%\end{tabular}
%\end{center}
%\end{table}
%\setlength{\tabcolsep}{1.4pt}

\section{Conclusions}
\label{sec:concl}

In this paper, we have presented a novel approach to incorporate geometric constraints modeled as polynomial equation systems into the higher-order graph matching framework. We have shown example formulations for three important geometric problems in computer vision and have shown the robustness of the approach to handle noise and outliers through extensive simulations. Finally, we have also shown that this framework allows us to handle correspondence problems with multiple motions using the same geometric constraints. 

In future work, we plan to look at the practical aspects of applying this higher-order formulation to geometric matching problems. In our formulation, building the affinity tensor involves computation of the resultant of a large number of small fixed size matrices (e.g. $8\times 8$ for the P3P problem). This is the most computationally intensive part of the algorithm but it is infinitely parallelizable. Therefore, we will look at distributing this computation using parallel GPU cores \cite{agullo2011qr,kruger2003linear} and that should allow us to tremendously scale down the tensor build time.

{\small
\bibliographystyle{ieee}
\bibliography{egbib}

\begin{thebibliography}{10}\itemsep=-1pt

\bibitem{agullo2011qr}
E.~Agullo, C.~Augonnet, J.~Dongarra, M.~Faverge, H.~Ltaief, S.~Thibault, and
  S.~Tomov.
\newblock Qr factorization on a multicore node enhanced with multiple gpu
  accelerators.
\newblock In {\em Parallel \& Distributed Processing Symposium (IPDPS), 2011
  IEEE International}, pages 932--943. IEEE, 2011.

\bibitem{cheng2013supermatching}
Z.~Cheng, Y.~Chen, R.~Martin, Y.~Lai, and A.~Wang.
\newblock Supermatching: Feature matching using supersymmetric geometric
  constraints.
\newblock {\em IEEE transactions on visualization and computer graphics}, 2013.

\bibitem{corless2004qr}
R.~M. Corless, S.~M. Watt, and L.~Zhi.
\newblock Qr factoring to compute the gcd of univariate approximate
  polynomials.
\newblock {\em Signal Processing, IEEE Transactions on}, 52(12):3394--3402,
  2004.

\bibitem{cour2006balanced}
T.~Cour, P.~Srinivasan, and J.~Shi.
\newblock Balanced graph matching.
\newblock In {\em NIPS}, volume~2, page~6, 2006.

\bibitem{duchenne2011tensor}
O.~Duchenne, F.~Bach, I.-S. Kweon, and J.~Ponce.
\newblock A tensor-based algorithm for high-order graph matching.
\newblock {\em Pattern Analysis and Machine Intelligence, IEEE Transactions
  on}, 33(12):2383--2395, 2011.

\bibitem{fischler1981random}
M.~A. Fischler and R.~C. Bolles.
\newblock Random sample consensus: a paradigm for model fitting with
  applications to image analysis and automated cartography.
\newblock {\em Communications of the ACM}, 24(6):381--395, 1981.

\bibitem{kruger2003linear}
J.~Kr{\"u}ger and R.~Westermann.
\newblock Linear algebra operators for gpu implementation of numerical
  algorithms.
\newblock In {\em ACM Transactions on Graphics (TOG)}, volume~22, pages
  908--916. ACM, 2003.

\bibitem{kukelova2011closed}
Z.~Kukelova, M.~Bujnak, and T.~Pajdla.
\newblock Closed-form solutions to minimal absolute pose problems with known
  vertical direction.
\newblock In {\em Computer vision--ACCV 2010}, pages 216--229. Springer, 2011.

\bibitem{laidacker1969another}
M.~Laidacker.
\newblock Another theorem relating sylvester's matrix and the greatest common
  divisor.
\newblock {\em Mathematics Magazine}, pages 126--128, 1969.

\bibitem{lee2011hyper}
J.~Lee, M.~Cho, and K.~M. Lee.
\newblock Hyper-graph matching via reweighted random walks.
\newblock In {\em Computer Vision and Pattern Recognition (CVPR), 2011 IEEE
  Conference on}, pages 1633--1640. IEEE, 2011.

\bibitem{leordeanu2005spectral}
M.~Leordeanu and M.~Hebert.
\newblock A spectral technique for correspondence problems using pairwise
  constraints.
\newblock In {\em Computer Vision, 2005. ICCV 2005. Tenth IEEE International
  Conference on}, volume~2, pages 1482--1489. IEEE, 2005.

\bibitem{naroditsky2012two}
O.~Naroditsky, X.~S. Zhou, J.~Gallier, S.~I. Roumeliotis, and K.~Daniilidis.
\newblock Two efficient solutions for visual odometry using directional
  correspondence.
\newblock {\em Pattern Analysis and Machine Intelligence, IEEE Transactions
  on}, 34(4):818--824, 2012.

\bibitem{ochs2012higher}
P.~Ochs and T.~Brox.
\newblock Higher order motion models and spectral clustering.
\newblock In {\em Computer Vision and Pattern Recognition (CVPR), 2012 IEEE
  Conference on}, pages 614--621. IEEE, 2012.

\bibitem{park2013fast}
S.~Park, S.-K. Park, and M.~Hebert.
\newblock Fast and scalable approximate spectral matching for higher-order
  graph matching.
\newblock {\em IEEE transactions on pattern analysis and machine intelligence},
  2013.

\bibitem{schellewald2005probabilistic}
C.~Schellewald and C.~Schn{\"o}rr.
\newblock Probabilistic subgraph matching based on convex relaxation.
\newblock In {\em Energy minimization methods in computer vision and pattern
  recognition}, pages 171--186. Springer, 2005.

\bibitem{zass2008probabilistic}
R.~Zass and A.~Shashua.
\newblock Probabilistic graph and hypergraph matching.
\newblock In {\em Computer Vision and Pattern Recognition, 2008. CVPR 2008.
  IEEE Conference on}, pages 1--8. IEEE, 2008.

\bibitem{zhou2012factorized}
F.~Zhou and F.~De~la Torre.
\newblock Factorized graph matching.
\newblock In {\em Computer Vision and Pattern Recognition (CVPR), 2012 IEEE
  Conference on}, pages 127--134. IEEE, 2012.

\bibitem{zhou2013deformable}
F.~Zhou and F.~De~la Torre.
\newblock Deformable graph matching.
\newblock In {\em Computer Vision and Pattern Recognition (CVPR), 2013 IEEE
  Conference on}, pages 2922--2929. IEEE, 2013.

\end{thebibliography}
}

\end{document}